\documentclass[letterpaper, 10pt, journal, twoside]{IEEEtran}

\IEEEoverridecommandlockouts % This command is only needed if
\makeatletter

\let\proof\@undefined
\let\endproof\@undefined
\makeatother

\usepackage{hyperref}
\hypersetup{
  colorlinks,
  citecolor=black,
  filecolor=black,
  linkcolor=black,
  urlcolor=black,
  pdfauthor={},
  pdfsubject={},
  pdftitle={}
}

\usepackage{graphicx}
\usepackage{setspace}
\usepackage{epstopdf}
\usepackage{float}
\usepackage{multirow,tabularx,makecell}
\usepackage{siunitx}
\usepackage{cellspace}

\usepackage{subcaption}
\usepackage{pifont}

\newcolumntype{D}{>{\hfill}N{3}{2}<{\hfill}}

\usepackage{algorithm}
\usepackage{etoolbox}\AtBeginEnvironment{algorithmic}{\scriptsize}
\usepackage[noend]{algpseudocode}

\usepackage{wrapfig}
\usepackage[font={small}]{caption}
\usepackage{scalefnt}

\usepackage[export]{adjustbox}
\usepackage{environ}
\usepackage[para, bottom]{footmisc}
\usepackage{color}
\usepackage[dvipsnames]{xcolor}
\usepackage{url}

\usepackage{amsmath}
\usepackage{amssymb,amsmath}
\usepackage{mathtools}
\usepackage{siunitx}
\usepackage[nameinlink, noabbrev, capitalize]{cleveref}
\usepackage{cite}

\usepackage{trimclip}

\usepackage{isotope}
\usepackage{listings}
\makeatletter
\def\lst@makecaption{%
  \def\@captype{table}%
  \@makecaption
}
\makeatother

\makeatletter
\def\BState{\State\hskip-\ALG@thistlm}
\makeatother

\definecolor{dark_green}{rgb}{0.0, 0.6, 0.0}

\newcommand{\notok}[1]{{\color{BrickRed}{#1}}}
\newcommand{\ok}[1]{{\color{ForestGreen}{#1}}}

\newcommand{\rev}[1]{{\color{black} {#1}}}

\newcommand{\PUBLISHEDIN}{IEEE Robotics and Automation Letters}
\newcommand{\DOI}{10.1109/LRA.2024.3390596} % you will not get a DOI until the paper is actually published, so update this when you get it and reupload the new preprint to all systems
\usepackage[placement=top,vshift=-7]{background}
\SetBgScale{1.0}
\SetBgContents{\PUBLISHEDIN. PREPRINT VERSION - DO NOT DISTRIBUTE. \href{https://doi.org/\DOI}{DOI \DOI}}
\SetBgColor{black}
\SetBgAngle{0}
\SetBgOpacity{1.0}

\usepackage{tikz}
\usepackage{pgfplots}
\usepgfplotslibrary{groupplots}
\usetikzlibrary{backgrounds,arrows,automata,shapes,positioning,calc,through}
\pgfplotsset{compat=newest}
\pgfdeclarelayer{background}
\pgfdeclarelayer{foreground}
\pgfsetlayers{background,main,foreground}

\tikzset{
  state/.style={
    rectangle,
    draw=black, very thick,
    minimum height=1.0em,
    text centered,
  },
  legend_box/.style={
    rectangle,
    draw=black,
    text centered,
  },
  normalstate/.style={
    rectangle,
    draw=black, very thick,
    minimum height=2.9em,
    minimum width=6.25em,
    text centered,
  },
  finalstate/.style={
    rectangle,
    double=white,
    double distance=0.1em,
    inner sep=0.2em,
    draw=black, very thick,
    minimum height=2.90em,
    minimum width=6.25em,
    text centered,
  },
  initialstate/.style={
    rectangle,
    double=white,
    double distance=0.1em,
    inner sep=0.2em,
    draw=black, very thick,
    minimum height=2.90em,
    minimum width=6.25em,
    text centered,
  },
  point/.style={
    circle,
    inner sep=0pt,
    minimum size=3pt,
    fill=red
  },
  adder/.style={
    circle,
    inner sep=2pt,
    minimum size=0.3in,
    draw=black, very thick,
    text centered
  },
  arrow/.style={
    thick,
    ->,
  >=stealth},
  darrow/.style={
    thick,
    <->,
  >=stealth},
  block/.style={
        draw,
        rectangle,
        rounded corners,
        inner sep=0pt,
        fill=white,
        fill opacity=1.0,
        text opacity=1.0
    }
}
\usepackage{soul}

\usepackage{scalerel}
\usetikzlibrary{svg.path}
\definecolor{orcidlogocol}{HTML}{A6CE39}
\tikzset{
  orcidlogo/.pic={
    \fill[orcidlogocol] svg{M256,128c0,70.7-57.3,128-128,128C57.3,256,0,198.7,0,128C0,57.3,57.3,0,128,0C198.7,0,256,57.3,256,128z};
    \fill[white] svg{M86.3,186.2H70.9V79.1h15.4v48.4V186.2z}
    svg{M108.9,79.1h41.6c39.6,0,57,28.3,57,53.6c0,27.5-21.5,53.6-56.8,53.6h-41.8V79.1z M124.3,172.4h24.5c34.9,0,42.9-26.5,42.9-39.7c0-21.5-13.7-39.7-43.7-39.7h-23.7V172.4z}
    svg{M88.7,56.8c0,5.5-4.5,10.1-10.1,10.1c-5.6,0-10.1-4.6-10.1-10.1c0-5.6,4.5-10.1,10.1-10.1C84.2,46.7,88.7,51.3,88.7,56.8z};
  }
}
\newcommand\orcidicon[1]{\href{https://orcid.org/#1}{\mbox{\scalerel*{
        \begin{tikzpicture}[yscale=-1,transform shape]
          \pic{orcidlogo};
        \end{tikzpicture}
}{|}}}}

\title{Fast Swarming of UAVs in GNSS-denied Feature-\rev{poor} Environments without \\ Explicit Communication}

\author{Ji\v{r}\'{i} Horyna$^{1\orcidicon{0000-0001-6614-0928}}$, V\'{i}t Kr\'{a}tk\'{y}$^{1\orcidicon{0000-0002-1914-742X}}$, V\'{a}clav Pritzl$^{1\orcidicon{0000-0002-7248-6666}}$,
Tom\'{a}\v{s} B\'{a}\v{c}a$^{1\orcidicon{0000-0001-9649-8277}}$,
Eliseo Ferrante$^{2\orcidicon{0000-0002-2213-8356}}$ and
Martin Saska$^{1\orcidicon{0000-0001-7106-3816}}$
  \thanks{Manuscript received: October 19, 2023; Revised February 21, 2024; Accepted March 22, 2024.}
  \thanks{
  This paper was recommended for publication by Editor M. Ani Hsieh upon evaluation of the Associate Editor and Reviewers' comments.
This work was funded by CTU grant no SGS23/177/OHK3/3T/13, by the Czech Science Foundation (GAČR) under research project no. 23-07517S, by the European Union under the project Robotics and advanced industrial production (reg. no. CZ.02.01.01/00/22\_008/0004590), and by the Technology Innovation Institute - Sole Proprietorship LLC, UAE, under the Research Project Con-
tract No. TII/ATM/2032/2020.
}
  \thanks{$^1$Multi-Robot Systems Group, Faculty of Electrical Engineering, Czech Technical University in Prague, Technicka 2, Prague, Czech Republic, {\tt\footnotesize\{\href{mailto:horynjir@fel.cvut.cz}{horynjir}|\href{mailto:vit.kratky@fel.cvut.cz}{kratkvit}|\href{mailto:pritzvac@fel.cvut.cz}{pritzvac}|\href{mailto:bacatoma@fel.cvut.cz}{bacatoma}|\href{mailto:martin.saska@fel.cvut.cz}{martin.saska}\}@fel.cvut.cz}.}
  \thanks{$^2$Department of Computer Science, Vrije Universiteit Amsterdam, Amsterdam, The Netherlands, {\tt\footnotesize\href{mailto:e.ferrante@vu.nl}{e.ferrante@vu.nl}}. 
  \newline 
  Digital Object Identifier (DOI): see top of this page.}
}

\begin{document}

\maketitle

%%{ ABSTRACT

\begin{abstract}
  A decentralized swarm approach for the fast cooperative flight of Unmanned Aerial Vehicles (UAVs) in feature-\rev{poor} environments without any external localization and communication is introduced in this paper.
  A novel model of a UAV neighborhood is proposed to achieve robust onboard mutual perception and flocking state feedback control, which is designed to decrease the inter-agent oscillations common in standard reactive swarm models employed in fast collective motion. 
  The novel swarming methodology is supplemented with an enhanced Multi-Robot State Estimation (MRSE) strategy to increase the reliability of the purely onboard localization, which may be unreliable in real environments.
  Although MRSE and the neighborhood model may rely on information exchange between agents, we introduce a communication-less version of the swarming framework based on estimating communicated states to decrease dependence on the often unreliable communication networks of large swarms.
  The proposed solution has been verified by a set of complex real-world experiments to demonstrate its overall capability in different conditions, including a UAV interception-motivated task with a group velocity reaching the physical limits of the individual hardware platforms.
\end{abstract}


%%{ MAIN TEXT

\vspace{-0.3cm}
\section{INTRODUCTION}
\begin{figure}[t]
  \setlength\belowcaptionskip{-1.4\baselineskip}
  \centering
  \includegraphics[width=0.47\textwidth]{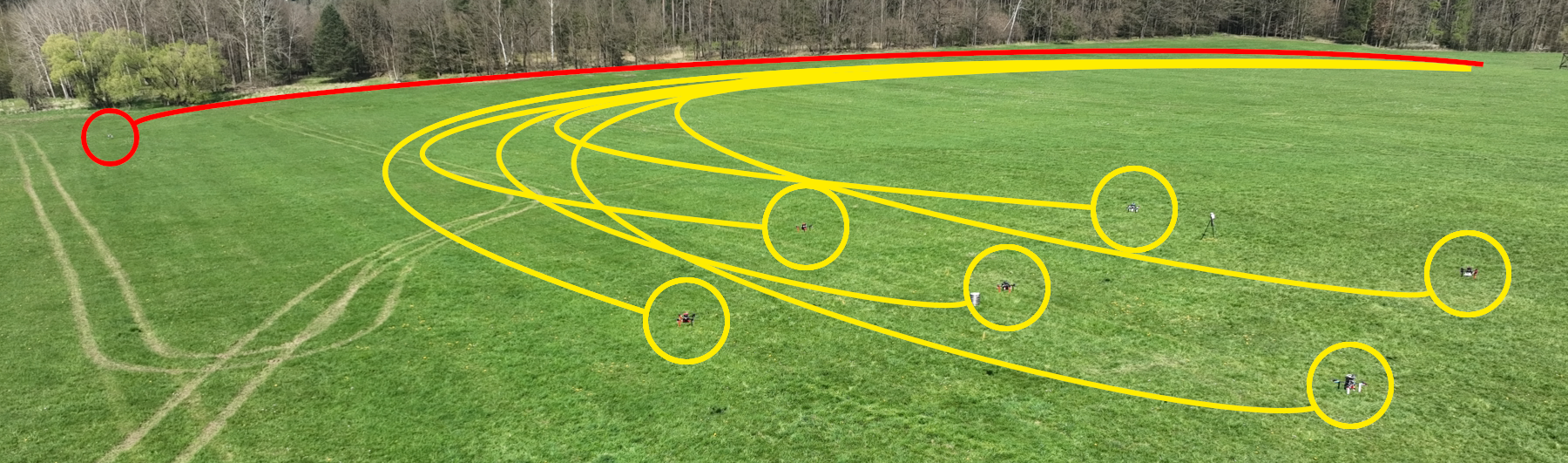}
  \caption{A swarm of six UAVs (yellow) tracking an intruder drone (red) using the proposed approach. The group velocity of \SI{5}{\meter\per\second} was reached, while the swarm stayed coherent without reliance on GNSS and communication. Circles show the positions of UAVs at the beginning of the experiment.}
  \label{fig:intro}
\end{figure}

\IEEEPARstart{A}{erial} robotic swarms could revolutionize various domains due to enhanced efficiency, robustness, fault tolerance, and cooperative problem-solving abilities. 
  These abilities are currently difficult to achieve due to the apparent complexity of the overall solution and the demanding system transfer from laboratories and simulations to the real world. 
  The available solutions \cite{olaronke2020swarmsreview} are limited mainly by the physical constraints and high demands on impractical sensory equipment, as well as the communication load. 
  To effectively deploy a swarm of UAVs (\autoref{fig:intro}) in tasks, such as exploration and monitoring~\cite{carpentiero2017swarm, albani2018dynamic} and search and rescue operations~\cite{cardona2019robot, couceiro2013collective}, it is necessary to mitigate the limitations mentioned. 

  The poor scalability of existing algorithms results in the inefficient achievement of swarm goals, especially concerning battery capacity limitations.
Increased group speed of the swarm is hard to achieve in the commonly used reactive flocking algorithms~\cite{hartman2006autonomous} due to coupling with inter-agent oscillations. These bottlenecks can be mitigated by task-suitable flocking control design and by using a proper filtration and estimation method for inputs of swarm control rules. 

  Following our experience~\cite{horyna2022sar, horyna2022estimation}, we infer that one of the elements for rapidly increasing the flexibility and versatility of a swarm in challenging conditions is to design a decentralized framework independent of an external infrastructure and relying exclusively upon onboard sensors. Such an approach makes the swarm robust against the interference, jamming, or spoofing of Global Navigation Satellite System (GNSS) signals, and allows for outdoor-indoor transition. On the other hand, the reliability of onboard state estimation needs to be ensured, since using visual or LiDAR localization methods is challenging in feature-\rev{poor} environments~\cite{horyna2022estimation}.

  Despite the theoretical abilities of modern communication modules, depending on the real-time information sharing among UAVs presents difficulties to swarm scalability, which may suffer from signal vulnerability in constrained environments, such as forests and urban areas. 
  Relying on none or only minimal communication~\cite{talamali2021less} makes a significant difference in the overall safety-critical robustness of the swarm. 

The proposed framework overcomes the common bottlenecks of swarm deployment, especially for the aspects of physical constraints, flexibility, reliability, dependence on external infrastructure, and communication. 
Our solution is designed for fast swarming in environments with insufficient amount of visual features (e.g., in plain grass fields, deserts, or long corridors~\cite{petracek2021caves}).
This is achieved by designing the swarming framework upon an enhanced Multi-Robot State Estimator (MRSE)~\cite{horyna2022estimation} of which the state estimate is adaptively fused with Visual Inertial Odometry (VIO).
Furthermore, UltraViolet Direction And Ranging (UVDAR)~\cite{uvdd2, walter2019uvdar} data are used to model the motion of a UAV's teammates as necessary for the proposed fast flocking control rules.

\vspace{-0.3cm}
\subsection{Related work}
Decentralized swarm architectures~\cite{olaronke2020swarmsreview} offer flexible and robust systems, crucial for reliably deploying multi-UAV teams in demanding real-world environments. Many of these architectures rely on global information sharing among robots for local flocking control \cite{viragh2014flocking, vasarhelyi2014flocking, vasarhelyi2018flocking}. Although these works analyze flocking models for swarm stability, they also lack real-world validation due to the difficulty of achieving a reliable communication network. In \cite{afzal2021Icra}, a bio-inspired control approach is proposed to navigate in obstacle-rich environments and prioritize UAV safety. The authors in \cite{petracek2020swarms} present a similar flocking model and analyze swarm stability concerning position estimation accuracy. In \cite{zhou2022swarm}, a swarm framework capable of cluttered environment navigation using exclusively onboard sensors was proposed. Another approach, inspired by the behavior of bees, is discussed in \cite{schmickl2011beeclust}. However, this solution is suitable for large groups of UAVs and dense environments. In \cite{schilling2018learning}, the authors demonstrate the training of convolutional neural networks for flocking using only raw camera images to generate velocity commands.
Compared to these works, the proposed approach is designed to overcome common physical constraints, especially the propagation speed of the swarm. The framework is resistant to \rev{a Single Point of Failure (SPoF). Specifically, it is resistant to} onboard state estimation failures coupled with a low amount of visual features in the environment.  

The literature covering the field of cooperative state estimation approaches is very limited. 
Research on cooperative state estimation and localization often focuses on a single UAV dealing with the loss of GNSS signal \mbox{\cite{goel2017distributed, qu2010cooperative1, qu2010cooperative2, qu2011cooperative, wan2014cooperative}}.
Existing approaches involve sharing GNSS, inertial, and Ultra-Wideband (UWB) data among UAVs for localization \cite{goel2017distributed}, re-localizing UAVs using known relative ranges \mbox{\cite{qu2011cooperative, qu2010cooperative1, qu2010cooperative2}}, or utilizing a leader-follower structure with GNSS-supported localization \cite{russell2019cooperative}. 
The work in~\cite{horyna2022estimation} enables the onboard cooperative estimation of lateral velocity directly integrated into a UAV control loop. 
Compared to the works above, the proposed enhanced MRSE does not require inter-agent communication, and allows for full lateral state estimation and adaptive fusion with a primary odometry source.

We demonstrate the capabilities of the proposed framework in experiments motivated by an important swarm application --- monitoring and dynamic object following~\cite {castrillo2022review}.
In~\cite{wang2020optimal}, a 3D swarm guidance law for this application is discussed, which uses a modified pure pursuit strategy for UAV swarm tracking and allows for formation maintenance and collision avoidance with the target. 
The authors of~\cite{sharma2022path} discussed limitations of several swarm algorithms deployed in an interception task, suggesting further exploration of this complex task.
A modular design of a swarm defense system primarily focused on formation control and management was proposed in~\cite{brust2021swarm}.
In contrast to these works, our solution was designed to respect the properties of real-world sensors and UAV dynamics, and was verified outside of laboratory conditions.

\vspace{-0.3cm}
\subsection{Contribution}
  The primary contribution of this work consists of designing a distributed flocking controller suitable for fast collective motion with added resistance to SPoF in ambiguous environments. The contributions are:
\begin{itemize}
\item The proposed distributed state feedback flocking controller allows to stabilize each UAV in an unambiguous position within the constellation and specify the desired collective velocity. 
As a result, the designed approach allows the swarm to fly at a significantly higher speed compared to existing approaches.
\item The incorporation of enhanced MRSE, which allows for safe deployment of the swarm in challenging conditions of environments with a low amount of visual features and no GNSS data available, where standard techniques of onboard localization fail.
\item The introduction of a swarming framework, including a neighborhood model representation for the precise and robust estimation and filtration of control inputs. 
\item The enhancement of MRSE by an adaptive fusion with the primary odometry source and a full-state representation, which we found to be crucial for decreasing estimated position degradation during agile maneuvers.  
\item Improvement of the robustness and flexibility of the proposed framework, including the enhanced MRSE, by proposing an approach for estimating immeasurable velocities of surrounding UAVs from observed swarming behavior, making communication an optional modality. 
\end{itemize}
To the best of our knowledge, this is the first framework proposing the distributed state estimation using only onboard sensors in a swarm of UAVs in ambiguous environments.

\begin{figure*}[ht]
  \setlength\abovecaptionskip{-0.4\baselineskip}
  \setlength\belowcaptionskip{-1.4\baselineskip}
  \centering
  \begin{subfigure}[t]{0.625\textwidth}
    \centering
    \includegraphics[page=1, trim={0.0cm 0.1cm 0.0cm 0.0cm}, width=1.0\textwidth, clip]{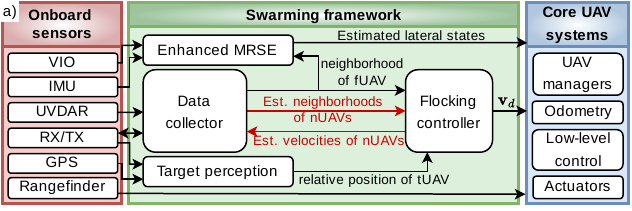}
    \label{fig:framework}
  \end{subfigure}
  \hfill
  \begin{subfigure}[t]{0.365\textwidth}
    \centering
    \includegraphics[page=1, trim={1.0cm 6.0cm 6.0cm 1.2cm}, width=1.0\textwidth, clip]{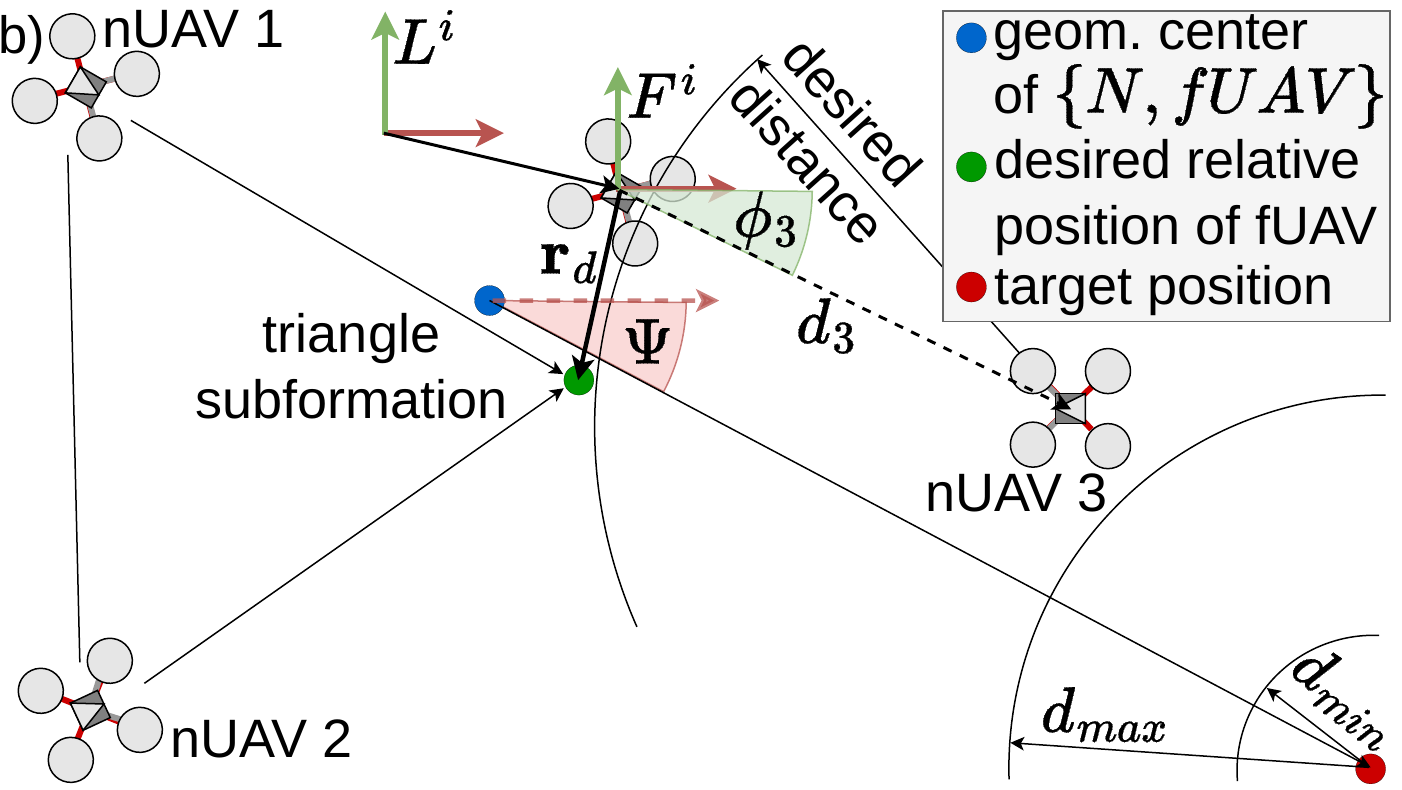}
    \label{fig:allinone}
  \end{subfigure}
  \caption{a) The system architecture for a single robot. High-level \emph{Flocking controller} determines the desired velocity of UAV based on the neighborhood state (modeled in \emph{Data collector}) and relative position of the goal (e.g., tUAV, in the interception-motivated task). The red arrows show the data flow in the process of the proposed neighbors' velocity estimation approach. In the \emph{Enhanced MRSE} block, VIO is fused with states of the enhanced cooperative estimator. b) Visualization of the desired position in the formation, group heading, geometrical formation rules, distance to target limits, and frames of reference.}
    \label{fig:framework_combo}
\end{figure*}

\vspace{-0.3cm}
\subsection{Problem statement \& Preliminaries}
The problem addressed in this paper regards a framework for fast UAV swarming in the challenging conditions of GNSS-denied environments lacking a significant amount of visual features. 
The proposed swarm framework relies on onboard sensors only, including 
a mutual perception system, a rangefinder for height estimation, a down-facing camera for VIO, an Inertial Measurement Unit (IMU), and an onboard computer.
We assume that the environment is unknown. 
In experiments where communication was allowed, we assume the UAVs can share lateral velocities among the UAVs in swarm to improve the model of neighboring UAVs.

The proposed swarm framework was designed upon the UVDAR smart sensor~\cite{uvdd2, walter2019uvdar}, and VINS-Mono~\cite{qin2018vins, BednarICUAS22_VIO}. The UVDAR system is an onboard visual mutual relative localization system using active UV markers and UV-sensitive cameras mounted on the frames of UAVs. It is resistant to erroneous detections and independent of external light sources. 
VINS-Mono integrates visual and inertial sensor measurements to estimate the camera pose in real-time. 
The MRSE~\cite{horyna2022estimation} is an approach for the distributed estimation of lateral velocities using visual observations of neighbors and communicated information. Herein, the enhanced MRSE is employed to increase the reliability of the onboard state estimation by fusion with the VINS-Mono of UAVs during fast flight in environments with a low amount of visual features.

Most of the mathematical expressions in this work are introduced in the local frame \emph{$L^i$}, which is fixed to the environment, aligned to the initial position of an \emph{i}-th UAV (see~\autoref{fig:framework_combo}), and oriented as an East-North-Up (ENU) coordinate system.
The fixed body frame \emph{$F^i$} is fixed to the UAV's center of mass, with all axes parallel to the axes of the local frame \emph{$L^i$}.
We assume parallel axes of reference frames of all UAVs thanks to the headings observable by magnetometers.
We introduce the terminology used in this paper as follows: 
\begin{itemize}
  \item Swarm ($S$) --- a set of homogeneous cooperating UAVs, 
  \item Focal UAV (fUAV) --- an arbitrary UAV from $S$, 
  \item Surroundings ($O$) --- the subset of $S$ containing UAVs observable by fUAV with the mutual perception system,
  \item Observable UAV (oUAV) --- an arbitrary UAV from $O$,
  \item Neighborhood ($N$) --- the subset of $O$ containing all UAVs that directly influence the behavior of the fUAV. $N$ is obtained using a selection algorithm~\cite{horyna2022sar} from $O$ filtered using the model described in \autoref{sec:model},  
  \item Neighbor (nUAV) --- an arbitrary UAV from $N$,
  \item Target (tUAV) --- an intruder UAV present in interception-motivated experiments that is not part of $S$.
\end{itemize}

\vspace{-0.2cm}
\section{Swarm framework}
The proposed swarm framework consists of several interconnected modules shown in the green area of \autoref{fig:framework_combo}. The 
\emph{Data collector} processes data about oUAVs and models their behavior using a bank of Linear Kalman Filters (LKFs) to estimate their state. 
This process plays a significant role in the framework pipeline since high-level control is influenced by the precision of the neighborhood model. 
Based on the model and the relative position of the goal, the intention of the fUAV is defined in \emph{Flocking controller}, where the desired velocity of fUAV is obtained using a state feedback controller. Reliability and performance of onboard VIO localization are enhanced in \emph{MRSE} with the observation of oUAVs. 
\emph{Target perception} emulates the relative position of tUAV in interception-motivated experiments and is replaceable with any sensory system measuring the relative position of tUAV.

\vspace{-0.3cm}
\subsection{Model and state estimation of surroundings}
\label{sec:model}
Reliable estimation of the oUAVs' position is crucial for robust swarm control, especially in the case of fast swarming. The onboard mutual perception systems used in real-world conditions (e.g., UVDAR) suffer from high noise and data dropouts due to uncertainties. 
Herein, oUAVs are modeled as point masses capable of motion in the horizontal plane. 
The high-level dynamics of \emph{i}-th oUAV is modeled as a discrete and decoupled Linear Time-Invariant (LTI) system:
\begin{align}
  \label{eq:model}
  \mathbf{x}^i_{k} =& \mathbf{A}\mathbf{x}^i_{k-1} + \mathbf{w}_k.
\end{align}
The state vector $\mathbf{x}^i_k$, state matrix $\mathbf{A}$, and process noise $\mathbf{w}_k$
are defined as:
\begin{align}
  \label{eq:model_info}
  \mathbf{x}_k^i = \left[
    \begin{smallmatrix}
      x \\
      y \\
      \dot{x} \\
      \dot{y} \\
      \ddot{x} \\
      \ddot{y} \\
    \end{smallmatrix}
    \right],
  \mathbf{A} = \left[
    \begin{smallmatrix}
      1 & 0 & \Delta t & 0 & \frac{\Delta t^2}{2} & 0 \\
      0 & 1 & 0 & \Delta t & 0 & \frac{\Delta t^2}{2} \\
      0 & 0 & 1 & 0 & \Delta t & 0 \\
      0 & 0 & 0 & 1 & 0 & \Delta t \\
      0 & 0 & 0 & 0 & 1 & 0 \\
      0 & 0 & 0 & 0 & 0 & 1 \\
    \end{smallmatrix}
    \right],
  \mathbf{w}_k \sim (\mathbf{0}, \mathbf{Q}), 
\end{align}
where $\Delta t$ is time between two consequent computational steps, and $\mathbf{Q}$ is the diagonal process covariance matrix with dimensions of matrix $\mathbf{A}$. The measurement model of the system is defined as $\mathbf{z}^i_{k} = \mathbf{H}\mathbf{x}^i_{k} + \mathbf{v}_k$, 
where $\mathbf{H}$ is a measurement matrix and $\mathbf{v}_k \sim (\mathbf{0}, \mathbf{R})$ is the measurement noise described by covariance matrix $\mathbf{R}$. Measurements of the states are obtained from two unsynchronized sources: the relative positions estimated by the onboard perception system transformed into \emph{L} frame, and either the communicated or estimated velocities of oUAVs. Thus, the measurement matrix is adaptive according to the measurement source: 
\begin{align}
  \label{eq:meas_matrix}
  \mathbf{H} = \left[
    \begin{smallmatrix}
      h_p & 0 & h_v & 0 & 0 & 0 \\
      0 & h_p & 0 & h_v & 0 & 0 \\
    \end{smallmatrix}
    \right]; h_p, h_v \in \{ 0, 1 \};h_p \neq h_v,
\end{align}
where $h_p = 1$,  when position measurement is obtained, and $h_v = 1$, when velocity measurement is obtained. The condition $h_p \neq h_v$ holds also for the rare case of measurements obtained at the same time, when corrections are applied sequentially.
The proposed state estimation approach uses one LKF for each oUAV. This approach forms a bank of LKFs (inbuilt in \emph{Data collector} from~\autoref{fig:framework_combo}), which makes the solution scalable and dynamic when compared to a single filter with states covering all oUAVs' states. The state estimate $\hat{\mathbf{x}}^i$ and state covariance $\mathbf{P}^i$ of \emph{i}-th filter are updated according to:
\begin{align}
  \label{eq:corr1}
  \mathbf{K}^i_k &= \mathbf{P}^i_{k|k-1} \mathbf{H}^T (\mathbf{H} \mathbf{P}^i_{k|k-1} \mathbf{H}^T + \mathbf{R})^{-1}, \\
  \label{eq:corr2}
  \hat{\mathbf{x}}^i_{k|k} &= \hat{\mathbf{x}}^i_{k|k-1} + \mathbf{K}^i_k (\mathbf{z}^i_k - \mathbf{H} \hat{\mathbf{x}}^i_{k|k-1}), \\
  \label{eq:corr3}
  \mathbf{P}^i_{k|k} &= \mathbf{P}^i_{k|k-1} - \mathbf{K}_k^i \mathbf{H} \mathbf{P}^i_{k|k-1}.
\end{align}
The prediction step of LKFs describes the evolution of the state estimate and state covariance in time using the model of the system with the previous state estimate and process noise:
\begin{align}
  \label{eq:pred}
  \hat{\mathbf{x}}^i_{k|k-1} &= \mathbf{A} \hat{\mathbf{x}}^i_{k-1|k-1}, \\
  \hat{\mathbf{P}}^i_{k|k-1} &= \mathbf{A}\mathbf{P}^i_{k-1|k-1}\mathbf{A}^T + \mathbf{Q}.
\end{align}
Values of matrices $\mathbf{Q}$ and $\mathbf{R}$ were set using the Normalized Estimation Error Squared (NEES) metric on the data obtained in the statistically processed real-world experiments in order to ensure consistency of the state estimator. 

\vspace{-0.3cm}
\subsection{Flocking control}
To achieve the requirements of fast collective motion behavior,
the proposed high-level flocking control law allows for specifying the desired group velocity via a feedforward component. The state feedback control provides local stabilization in an unambiguous position within the constellation. Such an approach is less dependent on the overall speed than traditional swarming algorithms, where individual control components may counteract~\cite{afzal2021Icra}. 
  We assume control of the UAV in the horizontal plane, taking only suitable nUAV candidates~\cite{horyna2022sar} into account, and using pre-filtered controller input variables (\autoref{sec:model}). The desired lateral velocity $\mathbf{v}_d$ is defined as 
\begin{equation}
  \label{eq:control}
  \begin{split}
    \mathbf{v}_d = &\overbrace{-k_p \mathbf{e}_p}^{\begin{tabular}{c}
      \tiny position feedback
    \end{tabular}} + \overbrace{-k_v \mathbf{e}_v}^{\begin{tabular}{c}
      \tiny velocity feedback
    \end{tabular}} + \overbrace{\mathbf{v}_G}^{\begin{tabular}{c}
      \tiny reference feedforward
    \end{tabular}},
  \end{split}
\end{equation}
where $\mathbf{e}_p = - \mathbf{r}_d$, $\mathbf{e}_v = - \mathbf{\dot{r}}_d$ are position and velocity control errors, respectively, $\mathbf{v}_G$ is the desired group velocity, and $k_p$, $k_v$ are position and velocity feedback gains, respectively. The control errors correspond to the negative relative state of the desired position in swarm formation. Using estimated bearings $\boldsymbol{\phi}$ and distances $\mathbf{d}$ of nUAVs, the desired position $\mathbf{r}_d$ (see~\autoref{fig:framework_combo}) is a function of the fUAV's neighborhood $N = N\{ \boldsymbol{\phi}, \mathbf{d} \}$ and group heading $\Psi$:
\begin{align}
  \label{eq:position_error}
  \mathbf{r}_{d} = f \left( N\{ \boldsymbol{\phi}, \mathbf{d} \}, \Psi \right) = \sum^{\left|N\right|}_i w \left( \phi_i, \Psi \right) g \left( \phi_i, d_i \right), %idk if ok
\end{align}
where $\phi_i, d_i$ is the bearing and distance of \emph{i}-th nUAV, respectively and the function $g \left( \phi_i, d_i \right)$ defines the desired position of the fUAV concerning the nUAV in frame \emph{$F^i$}, following the predefined geometrical rules drawn in~\autoref{fig:framework_combo}. The predefined geometrical rules intend to keep two close nUAVs in a triangle subformation with the fUAV, or keep a desired distance to the nUAV when triangle formation is inefficient (e.g., due to thresholding of an angle between bearings of nUAVs). The group heading $\Psi$ is determined as an angle between the x-axis of frame \emph{$L^i$} and the line intersecting the geometrical center of \emph{N} and swarm goal position. The weight function $w \left( \phi_i, \Psi \right)$ determines a weight coefficient with respect to angle difference $\theta_i = \left|\phi_i - \Psi\right|$ to fulfill the conditions
\begin{align}
  \label{eq:weight}
  \sum_i w \left( \phi_i, \Psi \right) = 1 \land w \left( \phi_i, \Psi \right) < w \left( \phi_j, \Psi \right) \forall \theta_i > \theta_j.
\end{align}
The weight coefficient is lower for the bearings of nUAVs further from the group heading. Thus, the weight function in \eqref{eq:position_error} decreases the influence of nUAVs flying behind the fUAV, which reduces inter-agent oscillations of swarm UAVs.

The desired group velocity $\mathbf{v}_G$ from \eqref{eq:control} is a feedforward part of the controller, and determines the forward motion of the swarm. To secure smooth deceleration of the swarm, $\mathbf{v}_G$ is scaled according to:
\begin{align}
  \label{eq:vg}
  \mathbf{v}_G =  
    \begin{cases}
      0 & \quad \text{if } \lVert\mathbf{r}_T\rVert \leq d_{min}, \\
      v_D & \quad \text{if } \lVert\mathbf{r}_T\rVert > d_{max}, \\
      v_D \frac{\lVert\mathbf{r}_T\rVert - d_{min}}{d_{max} - d_{min}} & \quad \text{otherwise},
    \end{cases}
\end{align}
where $v_D$ is the desired maximal group velocity of the swarm, $\mathbf{r}_T$ is the relative position of the target, $d_{min}$ is the lower-distance-to-target limit, and $d_{max}$ is the upper-distance-to-target limit. 
Since the goal swarm position can be substituted by tUAV in an interception-motivated task, the tUAV is added into a set of nUAVs once the fUAV is in the $d_{min}$ radius from the tUAV. This enables us to approach the tUAV with the prevention of collisions. 

\vspace{-0.3cm}
\subsection{Enhanced MRSE}
The proposed swarming framework is supplemented with an MRSE to support onboard state estimation, herein enhanced to fulfill the demands of fast swarming. Similarly to \cite{horyna2022estimation}, the focal UAV's lateral states are estimated using camera observations of nUAVs and data from the IMU. 
However, we herein introduce several contributions making MRSE applicable for fast flight.

\subsubsection{LKF neighborhood model}
The estimate of the position of nUAVs influences the precision of MRSE. Thus, we use the bank of LKFs introduced in~\autoref{sec:model} to improve the position of nUAVs being tracked. 

\subsubsection{Full state estimation}
\label{sec:mrse_lkf}
The fUAV lateral states $\mathbf{x}^f
= \left[
    \begin{smallmatrix}
      x^f & y^f & \dot{x}^f & \dot{y}^f & \ddot{x}^f & \ddot{y}^f
    \end{smallmatrix}
    \right]^T$ estimated by MRSE are obtained as the fusion of:
\begin{itemize}
  \item position of fUAV $\left[ x^U, y^U \right]$, as estimated by observation of nUAVs through the localization system concerning the modeled neighborhood, 
  \item lateral acceleration measured by IMU $\left[\ddot{x}^{I}, \ddot{y}^{I}\right]$,
  \item and the desired velocity $\mathbf{v}_d$ of the fUAV defined in \eqref{eq:control}.
\end{itemize}
Fusion is performed via LKF similar to the estimation method introduced in \ref{sec:model}. The fUAV is modeled as 
\begin{align}
  \label{eq:focal_model}
  \mathbf{x}^f_{k} =& \mathbf{A}\mathbf{x}^f_{k-1} + \mathbf{B}\mathbf{u}_k + \mathbf{w}_k,
  \mathbf{u}_k = \mathbf{v}_{d, k}, e_d = e^{-\frac{\Delta t}{\tau}}, \\ 
  \mathbf{A} =& \left[
    \begin{smallmatrix}
      1 & 0 & \Delta t & 0 & \frac{\Delta t^2}{2} & 0 \\
      0 & 1 & 0 & \Delta t & 0 & \frac{\Delta t^2}{2} \\
      0 & 0 & e_d & 0 & \Delta t & 0 \\
      0 & 0 & 0 & e_d & 0 & \Delta t \\
      0 & 0 & 0 & 0 & 1 & 0 \\
      0 & 0 & 0 & 0 & 0 & 1 \\
    \end{smallmatrix}
    \right],
  \mathbf{B} = \left[
    \begin{smallmatrix}
      0 & 0 \\
      0 & 0 \\
      1 - e_d & 0 \\
      0 & 1 - e_d \\
      0 & 0 \\
      0 & 0 \\
    \end{smallmatrix}
    \right],
\end{align}
where $\mathbf{B}$ is the input matrix, $\mathbf{u}_k$ is the input of the system, $\tau$ is the time constant, and $\mathbf{v}_{d, k}$ is the desired velocity of the fUAV at time step $k$.
The measurement model $\mathbf{z}^f_k = \mathbf{H}^f\mathbf{x}^f_k + \mathbf{v}_k$ is used with unsynchronized measurements $x^U, y^U, \ddot{x}^I, \ddot{y}^I$. The measurement matrix $\mathbf{H}^f$ is defined as: 
\begin{align}
  \label{eq:meas_matrix}
  \mathbf{H}^f = \left[
    \begin{smallmatrix}
      h_p & 0 & 0 & 0 & h_a & 0 \\
      0 & h_p & 0 & 0 & 0 & h_a \\
    \end{smallmatrix}
    \right], h_p, h_a \in \{ 0, 1 \}, h_p \neq h_a,
\end{align}
where $h_p = 1$ when the position measurement is obtained, and $h_a = 1$ when the acceleration measurement is obtained.
The prediction step of the fUAV's state estimate is defined as:
\begin{align}
  \label{eq:pred}
  \hat{\mathbf{x}}^f_{k|k-1} &= \mathbf{A} \hat{\mathbf{x}}^f_{k-1|k-1} + \mathbf{B}\mathbf{u}_k, \\
  \hat{\mathbf{P}}^f_{k|k-1} &= \mathbf{A}\mathbf{P}^f_{k-1|k-1}\mathbf{A}^T + \mathbf{Q}.
\end{align}
A correction step equivalent to equations~\eqref{eq:corr1} - \eqref{eq:corr3} is applied to update the state estimate and covariance. The full state estimate with LKF fusion decreases the position accuracy degradation compared to the pure velocity estimation in \cite{horyna2022estimation}.

\subsubsection{Adaptive fusion}
The VIO is fused with MRSE using an adaptive fusion parameter $\lambda \in \left[0, 1\right]$ to increase the reliability and robustness of the onboard state estimate. The adaptive $\lambda$ allows for dynamically changing confidence in the VIO when the VIO state is not reliable (e.g., rapid change of light conditions and low amount of features). It is defined as:
\begin{align}
  \label{eq:lambda}
  \lambda_{k} =&
    \begin{cases}
      \lambda_{k-1} - q\Delta t & \quad \text{if } \lambda_{k-1} > \lambda_e, \\
      \lambda_{k-1} + q\Delta t & \quad \text{if } \lambda_{k-1} < \lambda_e, \\
      \lambda_{k-1} & \quad \text{otherwise},
    \end{cases}
\end{align}
where $q > 0$ is a constant parameter defining the rate of change of $\lambda$ in time, and $\lambda_e$ is the estimated fusion parameter at a given time. We define $\lambda_e$ as a function of the VIO qualitative and quantitative states, specifically:
\begin{equation}
  \label{eq:lambdat}
  \begin{split}
    \lambda_e =& \overbrace{\frac{c_f}{C_f}}^{\begin{tabular}{c}
      \tiny quantitative coefficient
    \end{tabular}} . \overbrace{\frac{\sum_i^{c_f} t_i}{t_A C_f}}^{\begin{tabular}{c}
      \tiny qualitative coefficient
    \end{tabular}} = \frac{c_f\sum_i^{c_f} t_i }{t_A C^2_f},
  \end{split}
\end{equation}
where $c_f$ is the current number of visual features, $C_f$ is the maximal number of features (limited by the VINS-Mono algorithm configuration), $t_i$ is the time since the \emph{i}-th feature is tracked, and $t_A$ is the average time of the visual features being tracked in the camera image.
To maintain the system modularity, VIO is fused with MRSE outside the LKF described in Section \ref{sec:mrse_lkf}. The fusion of VIO and MRSE is performed in the \emph{$L^i$} frame according to:
\begin{align}
\begin{split}
  \label{eq:fusion}
    \mathbf{p}_{F,k} &= \mathbf{p}_{F, k-1} + \lambda_k \Delta \mathbf{p}_{V} + \left(1-\lambda_k\right) \Delta \mathbf{p}_{M}, \\
    \mathbf{v}_{F,k} &= \lambda_k \mathbf{v}_{V, k} + \left(1-\lambda_k\right) \mathbf{v}_{M, k}, \\
    \mathbf{a}_{F,k} &= \lambda_k \mathbf{a}_{V, k} + \left(1-\lambda_k\right) \mathbf{a}_{M, k}, \\
    \Delta \mathbf{p}_{V} &= \mathbf{p}_{V, k} - \mathbf{p}_{V, k-1}, \\
    \Delta \mathbf{p}_{M} &= \mathbf{p}_{M, k} - \mathbf{p}_{M, k-1}, 
\end{split}
\end{align}
where $\mathbf{p}_{S, k}, \mathbf{v}_{S, k}$, and $\mathbf{a}_{S, k}$ are the lateral position, velocity, and acceleration of the fUAV at time step $k$, with subscripts $S \in \{F,V,M\}$ corresponding to the fused state, state estimated by VIO, and state estimated by MRSE, respectively. Fused states are passed to the block \emph{Core UAV systems} (see \autoref{fig:framework_combo}) with a static measurement covariance matrix, as the covariance matrix of VINS-Mono is not defined. 

\subsection{Velocity estimation of observable UAVs}
\label{sec:velest}
In the proposed swarm framework, it is assumed that the fUAV is aware of the velocities of the oUAVs used in the surroundings model. The fundamental approach is to broadcast velocities over a communication channel. Since explicit communication may be the subject of numerous bottlenecks in large swarming systems, we introduce an oUAV velocity estimation method substituting inter-agent communication.

\begin{figure}[t]
  \vspace{0.1cm}
  \setlength\belowcaptionskip{-1.1\baselineskip}
  \centering
  \includegraphics[page=1, trim={0.0cm 0.25cm 0.0cm 0.0cm}, width=0.48\textwidth, clip]{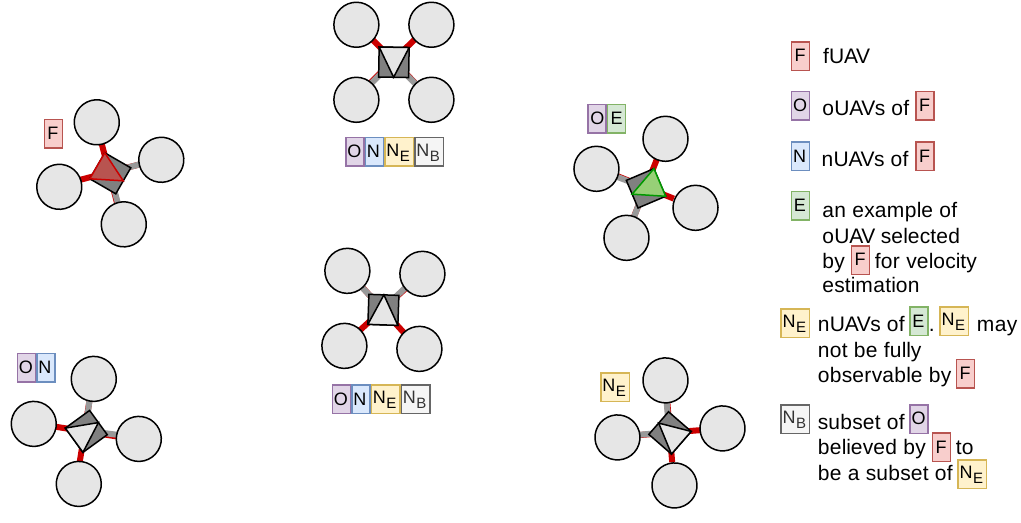}
  \caption{Example of an oUAV's neighborhood estimation. oUAV's neighborhood may not be fully observable by fUAV.}
  \label{fig:vel_est}
\end{figure}

We propose to estimate the desired velocity $\mathbf{v}^i_d$ of \emph{i}-th oUAV based on the state feedback control rule,~\eqref{eq:control}.
This approach assumes that the swarm is homogeneous and that all UAVs are under the same conditions resulting in the same intentions.
To estimate the neighborhood of an oUAV required in~\eqref{eq:control}, the fUAV selects a subset of UAVs from its surroundings \emph{O}, which is believed to be a subset of the \emph{i}-th oUAV neighborhood $N^i$. The estimation of the oUAV's neighborhood suffers high uncertainty, as the neighborhood of an oUAV may not be fully observable by fUAV (see~\autoref{fig:vel_est}). The desired velocity estimation error caused is mitigated when fused with the UVDAR observations (\autoref{sec:model}).
\begin{figure*}[ht]
  \vspace{0.1cm}
  \begin{subfigure}[t]{0.245\textwidth}
    \centering
    \includegraphics[width=1.0\textwidth]{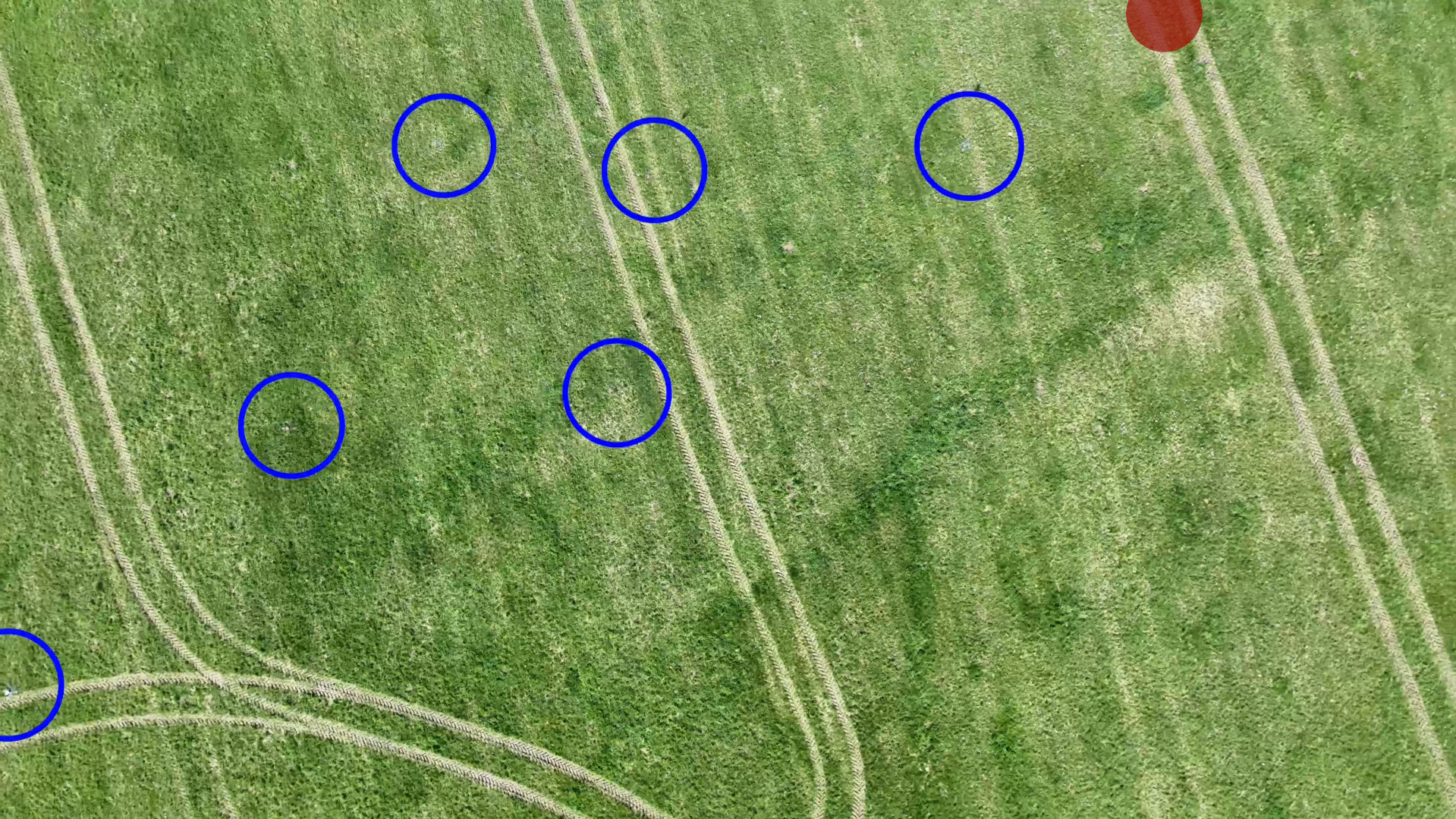}
    \caption{$t=t_0$}
    \label{fig:6a}
  \end{subfigure}
  \hfill
  \begin{subfigure}[t]{0.245\textwidth}
    \centering
    \includegraphics[width=1.0\textwidth]{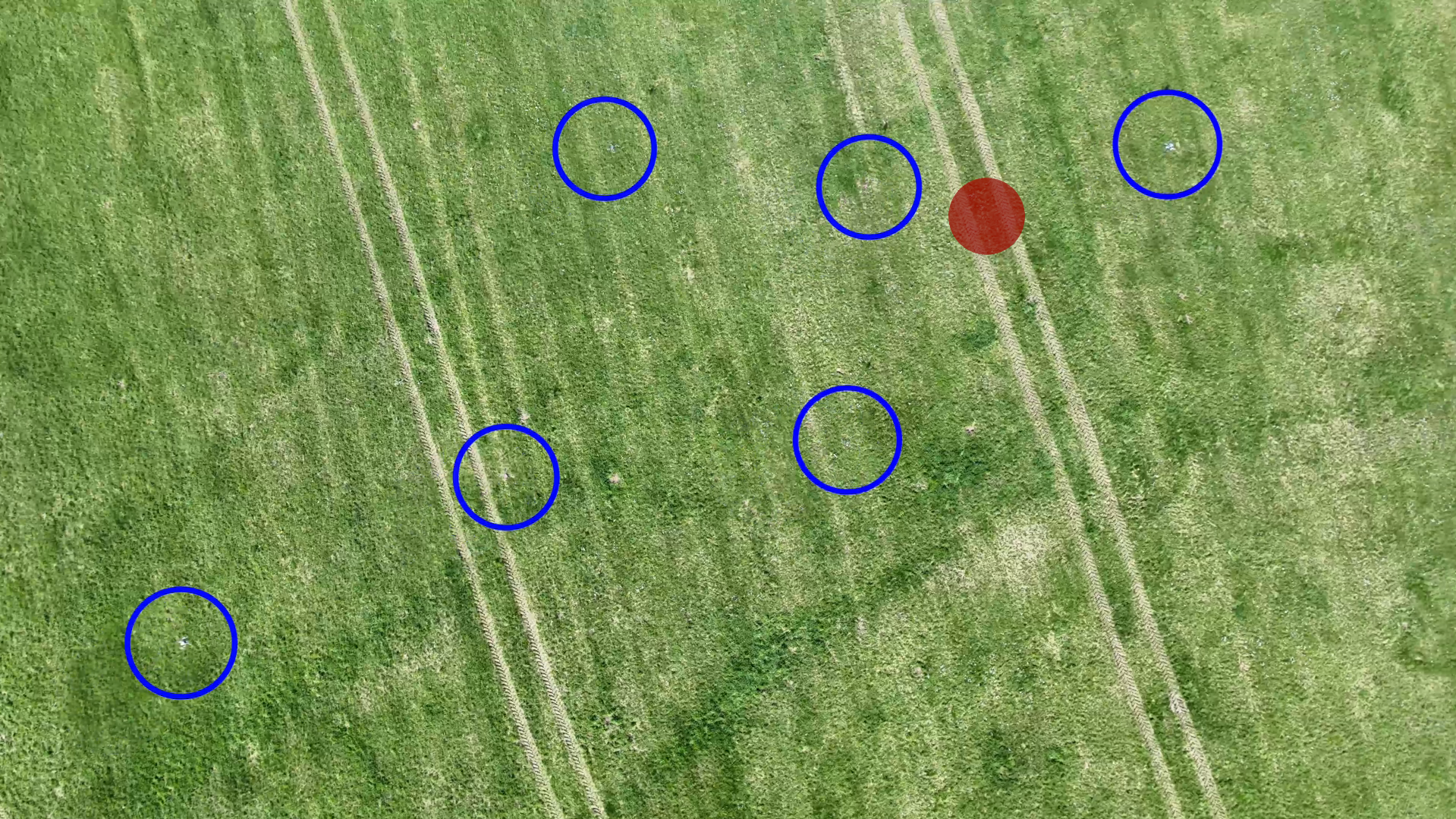}
    \caption{$t=t_0 + 5\,s$}
    \label{fig:6b}
  \end{subfigure}
  \hfill
  \begin{subfigure}[t]{0.245\textwidth}
    \centering
    \includegraphics[width=1.0\textwidth]{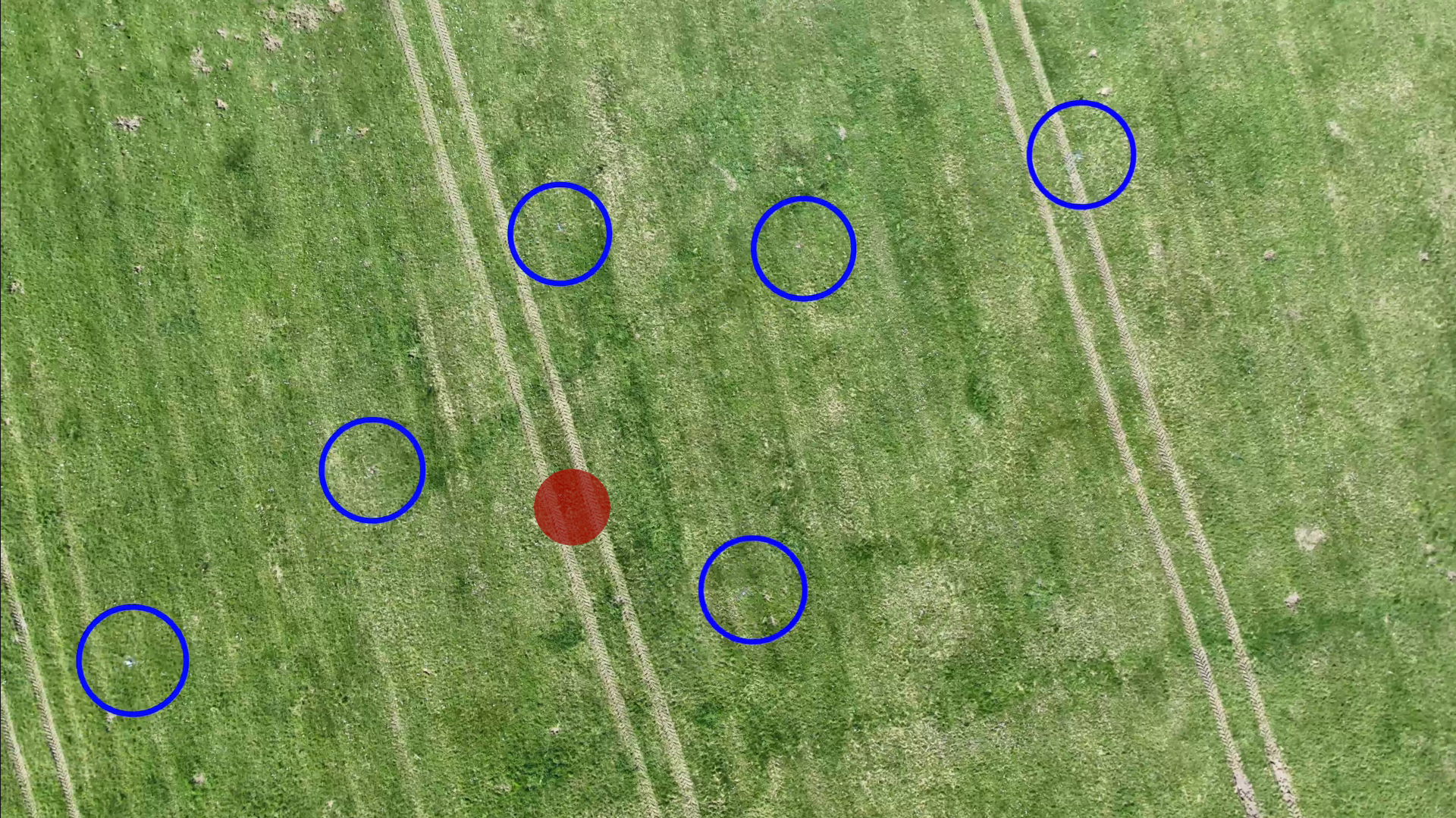}
    \caption{$t=t_0 + 10\,s$}
    \label{fig:6c}
  \end{subfigure}
  \hfill
  \begin{subfigure}[t]{0.245\textwidth}
    \centering
    \includegraphics[width=1.0\textwidth]{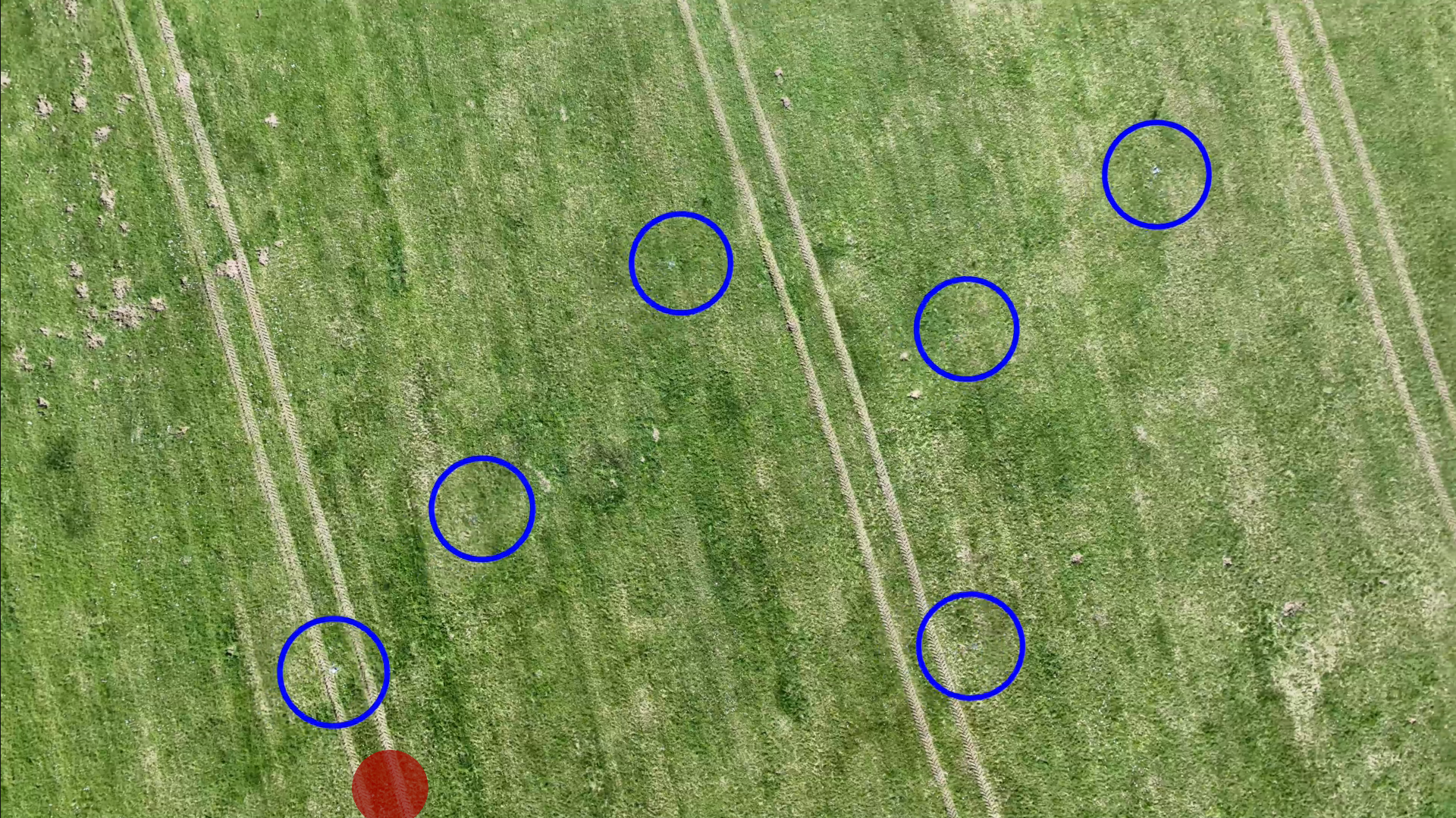}
    \caption{$t=t_0 + 15\,s$}
    \label{fig:6d}
  \end{subfigure}
  \caption{Top view of swarm approaching a static goal with group velocity \SI{5}{\meter\per\second}. The red circle is a motion reference. }
  \label{fig:sixuavs}
\end{figure*}

A discrete time-invariant linear state-space model was used for modeling the relation between estimated desired velocity $\mathbf{v}^i_d$ and estimated current velocity $\mathbf{v}^i_e$ of the \emph{i}-th oUAV:
\begin{align}
  \label{eq:ve}
  \mathbf{v}^i_e ( k+1 ) = q_1 \mathbf{v}^i_e ( k ) + q_2 \mathbf{v}^i_d ( k + 1 ).
\end{align}
To obtain parameters $q_1, q_2$, we formulate an overdetermined system of equations in the matrix form:
\begin{align}
\begin{pmatrix}
 \mathbf{v}^i_e (2) \\
 \mathbf{v}^i_e (3) \\
\vdots      \\
 \mathbf{v}^i_e (n) 
\end{pmatrix}
=
\begin{pmatrix}
 \mathbf{v}^i_e (1)   & \mathbf{v}^i_d (2) \\
 \mathbf{v}^i_e (2)   & \mathbf{v}^i_d (3) \\
\vdots        & \vdots \\
 \mathbf{v}^i_e (n-1) & \mathbf{v}^i_d (n)
\end{pmatrix}
\begin{pmatrix}
q_1 \\
q_2 
\end{pmatrix}.
\end{align}
The parameters are computed offline using the least square method on measured data for various velocity profiles. The overall estimation process is summarized in~Algorithm~1.

\begin{algorithm}
  \label{alg:velocity}
  \caption{Estimation of velocities of oUAVs}\label{alg:cap}
\begin{algorithmic}
  \Require $O^f, N^f, \mathbf{r}^f_t$ \Comment{Surroundings, neighborhood, relative target position}
  \Ensure $\mathbf{v}_e$ \Comment{List of estimated velocities of oUAVs}
    
  \State $\mathbf{v}_e = \left[ \ \right]$ 

  \For{$i \in O^f$}                    
    \State $N^i = \left[ \ \right]$
    \State {$N^i$ $\gets$ get\_nbhd({$O^f,\,i$}}) \Comment{Estimation of positions of \emph{i}-th oUAV's \emph{N}}
    \If{$O^f(i) \in N^f$} \Comment{Check if oUAV is nUAV}
      \State {$\mathbf{r}^i_f$ $=$ tf\_to\_nb($\left[0, 0\right]^T$,$\,i$)} \Comment{Relative position of fUAV in \emph{$F^i$} frame} 
      \State {$N^i$ $\gets$ $\mathbf{r}^i_f$} \Comment{Add position of fUAV to $N^i$}
    \EndIf
    \State $\mathbf{r}^i_t =$ tf\_to\_nb($\mathbf{r}^f_t,\,i$) \Comment{Transformation of relative target position to \emph{$F^i$} frame}
    \State $\mathbf{v}^i_d =$ estimate\_desired\_velocity($N^i, \mathbf{r}^i_t$) \Comment{\eqref{eq:control}}
    \State $\mathbf{v}^i_e =$ estimate\_current\_velocity($\mathbf{v}^i_d$) \Comment{\eqref{eq:ve}}
    \State $\mathbf{v}_e \gets \mathbf{v}^i_e$

  \EndFor
  \State \Return $\mathbf{v}_e$
\end{algorithmic}
\end{algorithm}

\vspace{-0.4cm}
\section{RESULTS}
% \vspace{-1.2cm}
\begin{figure}[b]
  \vspace{-0.4cm}
  \centering
  \includegraphics[page=1, trim={0.0cm 3.5cm 2.5cm 3.5cm}, width=0.48\textwidth, clip]{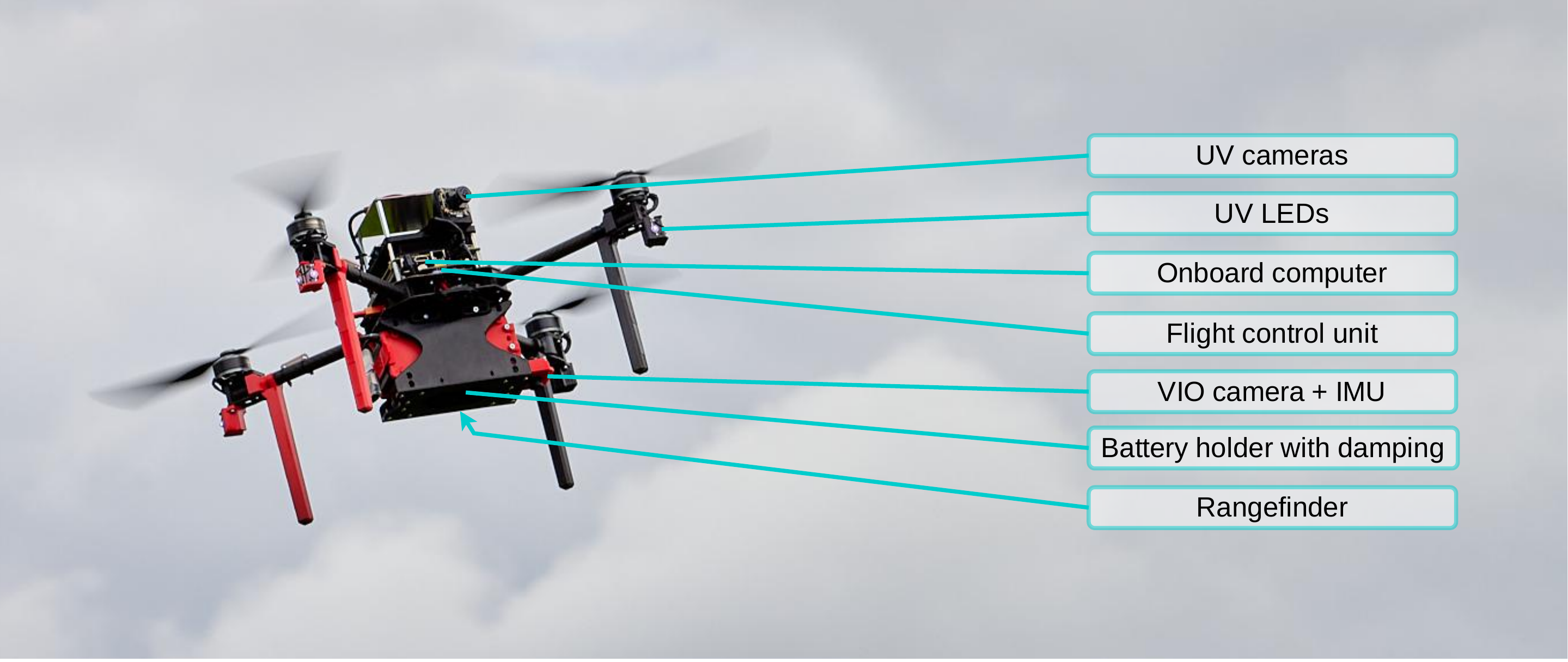}
  \caption{Hardware setup of UAVs used in experiments.}
  \label{fig:hw}
\end{figure}
After preliminary tests in realistic Gazebo simulations under Robot Operating System (ROS), the proposed swarming framework was validated through a series of real-world experiments. The experiments utilized customized quadrotor platforms (refer to \cite{HertJINTHW_paper} for more details) based on the Holybro X500 frame, equipped with a Pixhawk 4 flight control unit and an Intel NUC10i7FNK onboard computer (\autoref{fig:hw}).
In order to address the specific problem discussed in this study, each UAV was outfitted with a rangefinder to estimate the height above the ground, a damped downward-facing greyscale camera with wide-lens optics for VIO, and the UVDAR system. 
% The damped battery holder utilized the battery mass to reduce propeller vibrations acting on the VIO camera and IMU. 
The UVDAR system consisted of UV LEDs mounted on the UAV's arms and two UV-sensitive cameras. This configuration allowed for a horizontal field of view (FOV) of 320 degrees and a vertical FOV of 110 degrees.
To mitigate the blind spot at the back of the UAV, which spans 40 degrees, the heading control was actively employed to maximize the fUAV's observation of oUAVs. Therefore, employing a complete UVDAR setup with four cameras and no blind spots would yield similar outcomes.
The low-level control pipeline deployed on the UAVs is detailed in~\cite{baca2021mrs}.

We use the Cluster Velocity Ratio ($CVR$) and average distance to nUAV ($\overline{d}_n$) to evaluate the results of the introduced experiments. $CVR$ is determined as:
\begin{align}
  \label{eq:cvr}
  CVR = \frac{\lVert\dot{\mathbf{p}}_c\rVert}{v_{D}},
\end{align}
where $\mathbf{p}_c$ is the position of the geometrical center of the swarm. $CVR = 0$ when the swarm hovers, and $CVR = 1$ when the swarm approaches the target position with the maximum allowed group velocity. The average distance to the nUAV is defined as an average distance between all pairs of UAVs considered as neighbors. Both of the values, $CVR$ and $\overline{d}_n$, are obtained from GNSS data only for the purpose of evaluation. GNSS was not used for onboard processing, control, or state estimation in any of the introduced results. 
\begin{figure}[t]
  \setlength\belowcaptionskip{-1.4\baselineskip}
    \centering
    \includegraphics[page=1, trim={0.6cm 0.9cm 2.2cm 2.2cm}, width=0.48\textwidth, clip]{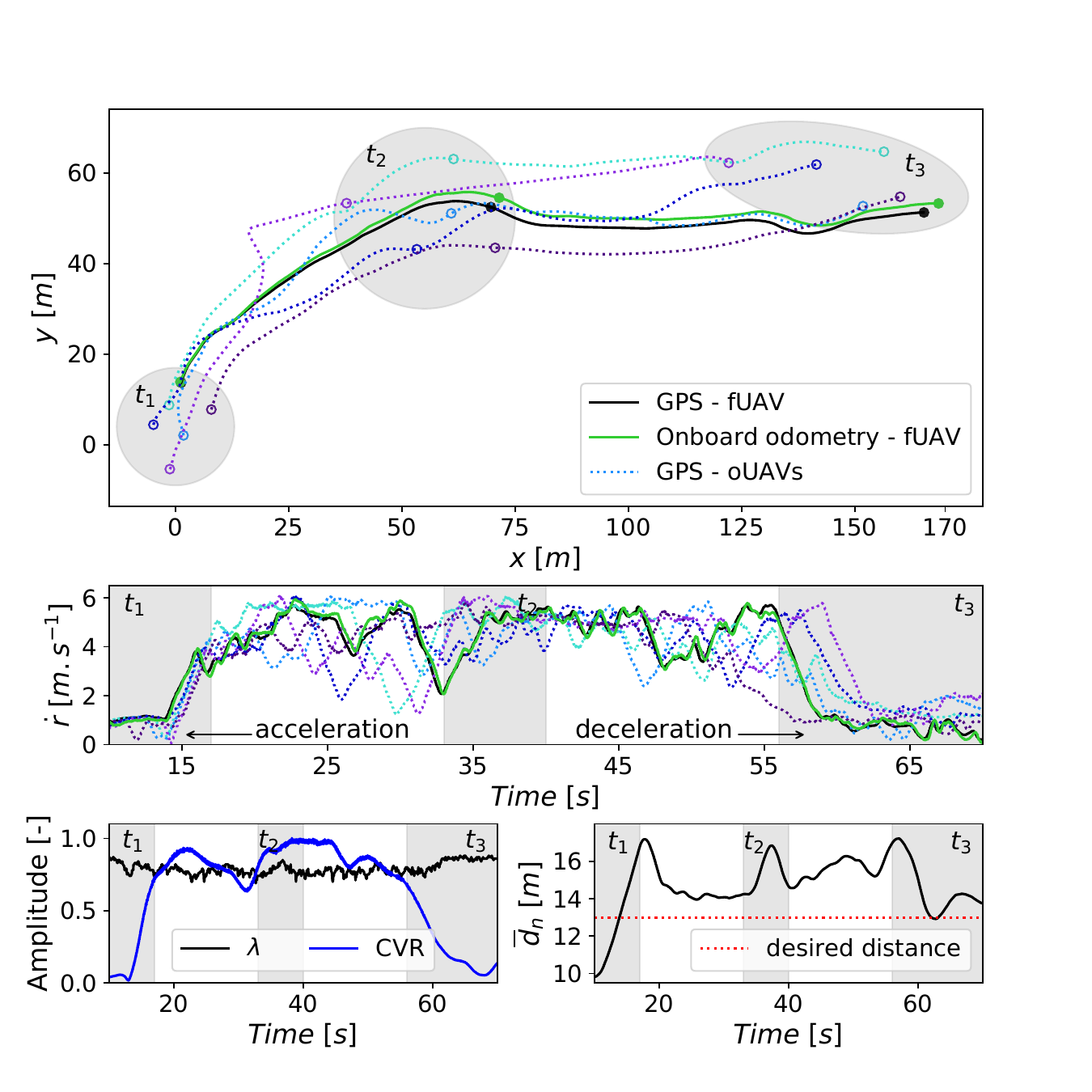}
    \caption{Swarm approaching a static goal (see~\autoref{fig:sixuavs} for top view images). Swarm UAVs' positions, velocities, and qualitative parameters are displayed. $\overline{d}_n =$~\SI{15,12}{\meter} with standard deviation \SI{5.07}{\meter}. GNSS is used as a ground truth, only.}
    \label{fig:single_data}
\end{figure} 

\vspace{-0.4cm}
\subsection{Real-world experiments}
The first set of experiments was performed with six UAVs (results from an experimental run is shown in \autoref{fig:sixuavs} and~\autoref{fig:single_data}) that were navigated towards a common goal using the proposed swarming framework. The UAVs used VIO fused with MRSE for self-localization and UVDAR for mutual perception. The swarm flew over \SI{200}{\meter} with an average group velocity reaching \SI{5,0}{\meter\per\second}. 
The fast movement of drones above the grass surface made tracking VIO features more challenging than slow flight. This led to decreased confidence in the VIO localization source and fusion with the MRSE using the coefficient $\lambda$. The lower $\lambda$ is, the lower the confidence put into the VIO.
The swarm reached the goal position without any collisions, having an average distance between the {nUAVs} of \SI{15,12}{\meter}, while the desired mutual distance was set to \SI{13,0}{\meter}. 
% The offset from the desired value is caused mainly by inaccuracy of UVDAR distance estimation. 
A quantitative comparison from multiple flights is summarized in~\autoref{tab:rw_res}. The noticeable standard deviation and an excessive average distance to nUAV are primarily caused by naturally high inaccuracy in distance estimation of the UVDAR sensor \cite{uvdd2}, which is caused by insufficient distance information in the UV camera image. This is also affected by camera calibration, as well as variable and asymmetric lighting conditions. The video from the experiment is available at \url{https://mrs.felk.cvut.cz/ral-2023-demo}.

\begin{table}[b]
  \setlength\belowcaptionskip{-1.0\baselineskip}
  \footnotesize
  \caption{Quantitative comparison from multiple flights of the first experiment. An average distance to nUAV $\overline{d}_n$, standard deviation $\sigma_d$, average $CVR$, and average and minimal $\lambda$ are shown.}
  \centering
  \setlength{\tabcolsep}{0.7em} % for the horizontal padding
{\renewcommand{\arraystretch}{1.25}% for the vertical padding
  \begin{tabular}{c|cccccc}
    \textbf{Flight No.} & 1 & 2 & 3 & 4 & 5 & 6 \\ \hline
    $\overline{d_n} \left[ m \right]$ & $15.12$ & $\rev{\mathbf{14.28}}$ & $14.93$  & $15.87$ & $15.42$ & $14.97$ \\
    $\sigma_d \left[ m \right]$   & $5.07$  & $4.85$ & $4.93$  & $5.03$  & $5.48$ & $\rev{\mathbf{4.62}}$ \\
    $\overline{CVR}$   & $0.89$ & $0.92$ & $0.87$ & $0.9$ & $\rev{\mathbf{0.93}}$ & $0.87$ \\
    $\overline{\lambda}$   & $0.82$ & $0.73$ & $0.82$ & $0.86$ & $0.81$ & $0.83$ \\
    $\lambda_{min}$   & $0.63$ & $0.39$ & $0.57$ & $0.54$ & $0.64$ & $0.37$ \\
  \end{tabular}}
  \label{tab:rw_res}
\end{table}

\begin{table}[b]
  \footnotesize
  \setlength\belowcaptionskip{-1.0\baselineskip}
  \caption{Quantitative comparison of the proposed framework with and without communication from multiple simulation flights. An average distance to nUAV $\overline{d}_n$, standard deviation $\sigma_d$, and the use of inter-agent communication are shown.}
  \centering
  \setlength{\tabcolsep}{0.3em} % for the horizontal padding
{\renewcommand{\arraystretch}{1.25}% for the vertical padding
  \begin{tabular}{c|cccccccc}
    \textbf{Flight No.} & 1 & 2 & 3 & 4 & 5 & 6 & 7 & 8 \\ \hline
    $\overline{d}_n \left[ m \right]$ & $13.78$ & $13.68$ & $\rev{\mathbf{13.34}}$  & $14.07$ & $15.75$ & $15.42$ & $15.91$ & $15.67$  \\
    $\sigma_d \left[ m \right]$   & $\rev{\mathbf{2.29}}$  & $2.34$ & $2.48$  & $2.37$  & $3.72$ & $3.58$ & $4.03$ & $3.82$ \\
    comm.   & \ding{51} & \ding{51} & \ding{51} & \ding{51} & \ding{55} & \ding{55} & \ding{55} & \ding{55}  \\
  \end{tabular}}
  \label{tab:sim_res}
\end{table}

\begin{figure}[t]
  \vspace{0.1cm}
  \setlength\belowcaptionskip{-1.3\baselineskip}
  \centering
  \includegraphics[page=1, trim={0.3cm 0.9cm 2.2cm 2.2cm}, width=0.47\textwidth, clip]{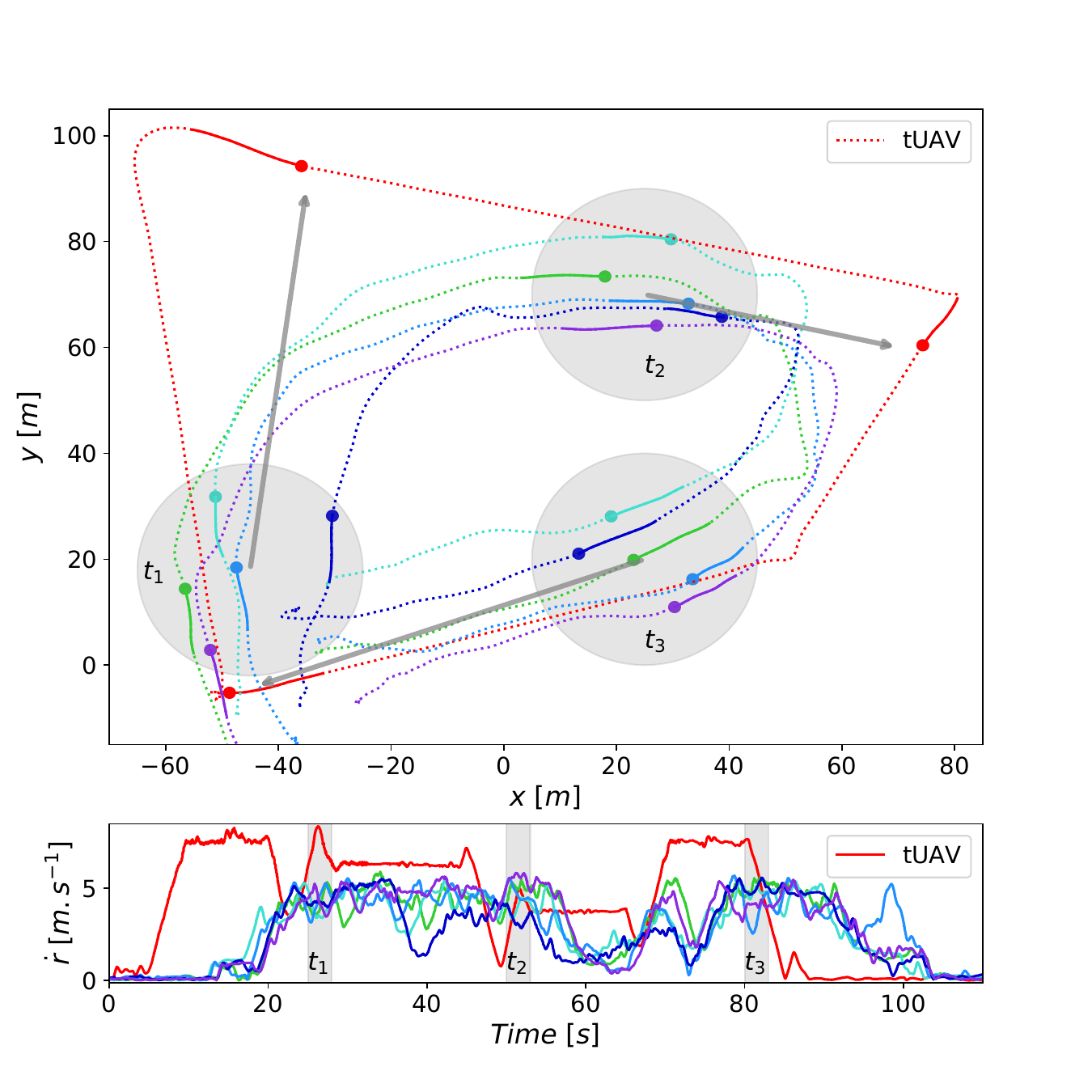}
  \caption{Swarm following a dynamically moving tUAV (red). The start and end position of the swarm is in the bottom left corner of the first graph. Grey areas show time intervals corresponding to marked positions from the upper graph.}
  \label{fig:interception}
\end{figure}

In the second set of real-world experiments (\autoref{fig:interception} and \autoref{fig:intro}), a swarm of five UAVs was supposed to follow a tUAV. Only onboard self-localization and mutual localization were used. Nevertheless, GNSS was deployed to emulate the onboard perception of swarm UAVs, which estimate the relative position of the tUAV. Similar results as the experiment in~\autoref{fig:single_data} were achieved. The proposed framework was able to control the team to follow dynamically moving objects and change flight direction without collisions or loss of cohesion.

\vspace{-0.3cm}
\subsection{Comparison of the proposed framework with and without communication}
\begin{figure}[t]
  \setlength\belowcaptionskip{-0.5\baselineskip}
  \centering
  \includegraphics[page=1, trim={1.5cm 2.5cm 2.0cm 0.7cm}, width=0.49\textwidth, clip]{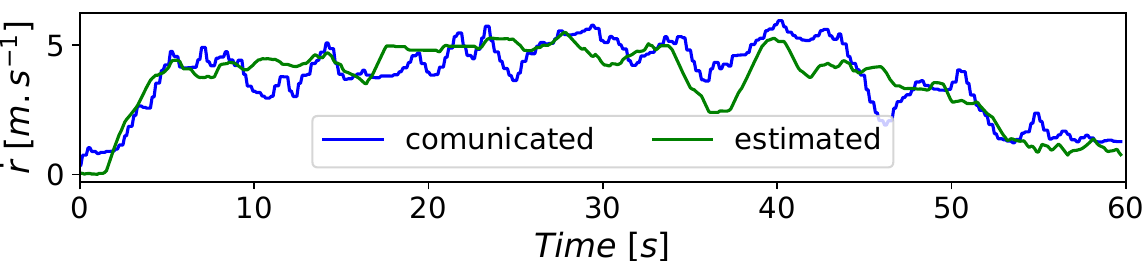}
  \caption{Comparison of the nUAV's velocity estimated by the fUAV and nUAV's onboard estimate.}
  \label{fig:velest}
\end{figure}

\begin{figure}[t]
  \setlength\belowcaptionskip{-1.0\baselineskip}
  \centering
  \includegraphics[page=1, trim={0.5cm 7.7cm 2.0cm 1.7cm}, width=0.49\textwidth, clip]{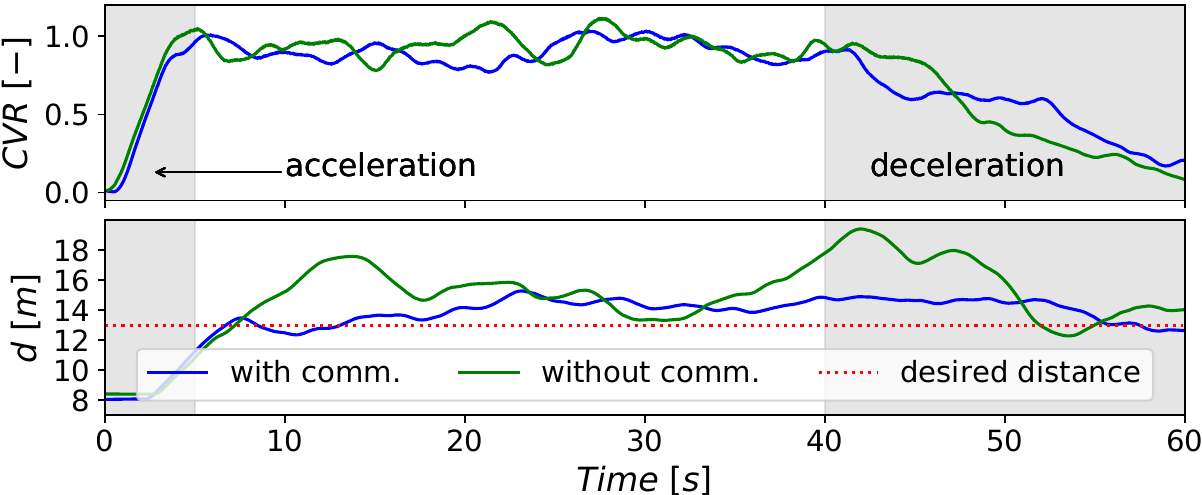}
  \caption{Comparison of the proposed framework, with and without communication. The inaccuracy of the velocity estimation method influences the shape of the formation due to the decrease in the accuracy of the neighborhood model. Results: $\overline{d}_n =$~\SI{13.78}{\meter}, $\sigma_d =$~\SI{2.29}{\meter} with communication, and $\overline{d}_n =$~\SI{15.75}{\meter}, $\sigma_d = $~\SI{3.72}{\meter} without communication.}
  \label{fig:sim_cvr}
\end{figure}

After the experimental campaign, the proposed framework, with and without inter-agent communication, was compared in the realistic Gazebo simulator as an additional demonstration of the framework's capabilities.
The swarm was commanded to fly to a desired place, similar to the real-world experiment in~\autoref{fig:single_data}. \autoref{fig:velest} shows a velocity profile of an nUAV estimated by the fUAV using an approach from~\autoref{sec:velest} compared to the velocity measured onboard the nUAV. 
The outcome of the proposed estimation approach is demonstrated in \autoref{fig:sim_cvr}. The $CVR$ value shows that the inaccuracy in mutual velocity estimation does not significantly impact the overall group speed. On the other hand, the proposed communication-less method influences the shape of the swarm, as it makes the neighborhood model less accurate. The quantitative comparison of the average mutual distances between nUAVs with standard deviations for several flights is shown in \autoref{tab:sim_res}. The average mutual distances have a lower standard deviation, which is caused by less challenging simulation conditions compared to real-world experiments.

\begin{table}[b]
  \footnotesize
  \setlength\belowcaptionskip{-1.0\baselineskip}
  \caption{Comparison of swarm features in related papers and our approach: \emph{included} (\ok{\ding{51}}) and \emph{not included} (\notok{\ding{55}}).}
  \centering
  \setlength{\tabcolsep}{0.7em} % for the horizontal padding
{\renewcommand{\arraystretch}{1.25}% for the vertical padding
  \begin{tabular}{c|cccccc}
    \textbf{Ref.} & \shortstack{Fast-flight \\ stability} & \shortstack{SPoF \\ resistance} & \shortstack{Comm.-loss \\ resistance} & \shortstack{GNSS \\ indep.} & \shortstack{Real-world \\ verification} \\ \hline
    \cite{horyna2022estimation} & \notok{\ding{55}} & \ok{\ding{51}} & \notok{\ding{55}} & \ok{\ding{51}} & \ok{\ding{51}} \\
    \cite{vasarhelyi2014flocking} & \ok{\ding{51}}  & \notok{\ding{55}} & \ok{\ding{51}}  & \notok{\ding{55}}  & \ok{\ding{51}} \\
    \cite{zhou2022swarm} & \ok{\ding{51}} & \notok{\ding{55}} & \notok{\ding{55}} & \ok{\ding{51}} & \ok{\ding{51}} \\
    \cite{schilling2018learning} & \notok{\ding{55}}  & \notok{\ding{55}} & \ok{\ding{51}}  & \ok{\ding{51}}  & \notok{\ding{55}} \\
    \cite{russell2019cooperative} & \notok{\ding{55}}  & \ok{\ding{51}} & \notok{\ding{55}}  & \notok{\ding{55}}  & \notok{\ding{55}} \\
    Ours & \ok{\ding{51}} & \ok{\ding{51}} & \ok{\ding{51}} & \ok{\ding{51}} & \ok{\ding{51}} \\
  \end{tabular}}
  \label{tab:related}
\end{table}

\vspace{-0.3cm}
\subsection{Comparison of the results with related works}
  Considering onboard localization only, the related works offer a narrow set of comparable approaches. We found~\cite{zhou2022swarm} to be the most related paper, where the authors introduced a swarm system capable of relatively fast coordinated motion through a cluttered environment. 
In comparison with this approach, we reached around a five times higher group velocity without the need for inter-agent communication.
The condition of featureless environments, which is barely touched on by state-of-the-art approaches, makes a fair comparison even more difficult. 
We provide a qualitative comparison of the key features of the proposed approach with the most relevant related works in~\autoref{tab:related}.

We compare the SPoF resistance demonstrated in~\autoref{fig:single_data} with our work~\cite{horyna2022estimation}.
\autoref{tab:mrse_res} shows three runs of both approaches, including the group velocity, trajectory length, velocity and position errors, and the number of swarm UAVs. Despite the significant increase in the trajectory length and velocity of UAVs in the proposed approach, the level of position and velocity errors of exclusive MRSE-aided onboard state estimation preserved in a feature-poor environment.  

\begin{table}[t]
  \vspace{0.2cm}
  \footnotesize
  \caption{Comparison of the proposed approach and our work~\cite{horyna2022estimation}. Group velocities ($v_g$), flight trajectory lengths ($l_t$), mean velocity errors ($v_e$), final position errors ($p_e$), and number of UAVs ($\left| S \right|$) are summarized.}
  \centering
  \setlength{\tabcolsep}{0.7em} % for the horizontal padding
{\renewcommand{\arraystretch}{1.25}% for the vertical padding
  \begin{tabular}{l|ccc|ccc}
    \textbf{Approach} & \multicolumn{3}{c|}{Enhanced MRSE} & \multicolumn{3}{c}{MRSE~\cite{horyna2022estimation}}\\ 
    \textbf{Flight No.} & 1 & 2 & 3 & 1 & 2 & 3 \\ \hline
    $v_g$ $\left[ \frac{m}{s} \right]$ & $\rev{\mathbf{5.08}}$ & $4.85$ & $4.62$  & $0.44$ & $0.46$ & $0.55$ \\
    $l_t$ $\left[ m \right]$   & $\rev{\mathbf{205}}$  & $182$ & $196$  & $29$  & $51$ & $42$ \\
    $v_e$ $\left[ \frac{m}{s} \right]$   & $0.15$ & $0.12$ & $0.14$ & $\rev{\mathbf{0.09}}$ & $0.10$ & $0.13$ \\
    $p_e$ $\left[ m \right]$   & $4.8$ & $3.5$ & $3.4$ & $\rev{\mathbf{1.1}}$ & $1.6$ & $1.2$ \\
    $\left|S\right|$ $\left[ - \right]$   & $5$ & $5$ & $\rev{\mathbf{6}}$ & $3$ & $3$ & $3$ 
  \end{tabular}}
  \label{tab:mrse_res}
\end{table}

\vspace{-0.3cm}
\section{CONCLUSION}
This paper presents a framework for a swarm of UAVs designed for fast collective motion in GNSS-denied applications. 
The framework incorporates a high-level state feedback control approach to guide agents towards a common objective. 
Onboard state estimation is supported by MRSE, which enhances VIO reliability during agile flight with full-state estimation, improved neighborhood modeling, and adaptive fusion. 
Additionally, to demonstrate the framework's capability of operating without inter-agent communication, a method for estimating the immeasurable velocities of neighboring UAVs is introduced.
Compared to existing approaches, this method demonstrates superior performance in terms of speed and, more importantly, reliability, which is crucial for deploying compact multi-robot aerial systems in real-world missions, where spoofing, jamming, or interference of GNSS signal may always happen, with fatal consequences for most of the state-of-the-art multi-UAV approaches.

%%}

%%{ THE BIBLIOGRAPHY
\bibliographystyle{IEEEtran}
\bibliography{bibliography.bib}
%%}

\end{document}